\documentclass{article} 
\usepackage{iclr2026_conference,times}

\usepackage{amsmath,amsfonts,bm}

\def\eqref#1{equation~\ref{#1}}

\def\1{\bm{1}}

\DeclareMathAlphabet{\mathsfit}{\encodingdefault}{\sfdefault}{m}{sl}
\SetMathAlphabet{\mathsfit}{bold}{\encodingdefault}{\sfdefault}{bx}{n}

\usepackage{hyperref}
\usepackage{url}
\usepackage{graphicx}
\usepackage{float}

\usepackage{multirow}
\usepackage{booktabs}
\usepackage{xcolor}
\usepackage{array}
\usepackage{wrapfig}
\usepackage{bbm}

\usepackage{authblk}

\title{Contextual Inference from Single Objects in Vision-Language Models}

\iclrfinalcopy 

\begin{document}
\fancypagestyle{plain}{
  \fancyhf{}
  \fancyfoot[C]{\thepage}
}
\thispagestyle{plain}
\pagestyle{plain}

\author{%
\textbf{Martina G. Vilas}\textsuperscript{1}\thanks{Equal contribution}\quad
\textbf{Timothy Schaumlöffel}\textsuperscript{1,2*}\quad
\textbf{Gemma Roig}\textsuperscript{1,2}\par
\par
\vspace{0.6em}
\textsuperscript{1}Goethe University Frankfurt\quad
\textsuperscript{2}The Hessian Center for AI\quad
}

\maketitle

\begin{abstract}
How much scene context a single object carries is a well-studied question in human scene perception, yet how this capacity is organized in vision-language models (VLMs)s remains poorly understood, with direct implications for the robustness of these models.
We investigate this question through a systematic behavioral and mechanistic analysis of contextual inference from single objects. 
Presenting VLMs with single objects on masked backgrounds, we probe their ability to infer both fine-grained scene category and coarse superordinate context (indoor vs. outdoor).
We found that single objects support above-chance inference at both levels, with performance modulated by the same object properties that predict human scene categorization.
Object identity, scene, and superordinate predictions are partially dissociable: accurate inference at one level neither requires nor guarantees accurate inference at the others, and the degree of coupling differs markedly across models.
Mechanistically, object representations that remain stable when background context is removed are more predictive of successful contextual inference. Scene and superordinate schemas are grounded in fundamentally different ways: scene identity is encoded in image tokens throughout the network, while superordinate information emerges only late or not at all. 
Together, these results reveal that the organization of contextual inference in VLMs is more complex than accuracy alone suggests, with behavioral and mechanistic signatures dissociating in ways that differ substantially across models. 
\end{abstract}

\section{Introduction}

\begin{wrapfigure}{r}{0.6\textwidth}
  \vspace{-10pt}
  \centering
  \includegraphics[width=0.58\textwidth]{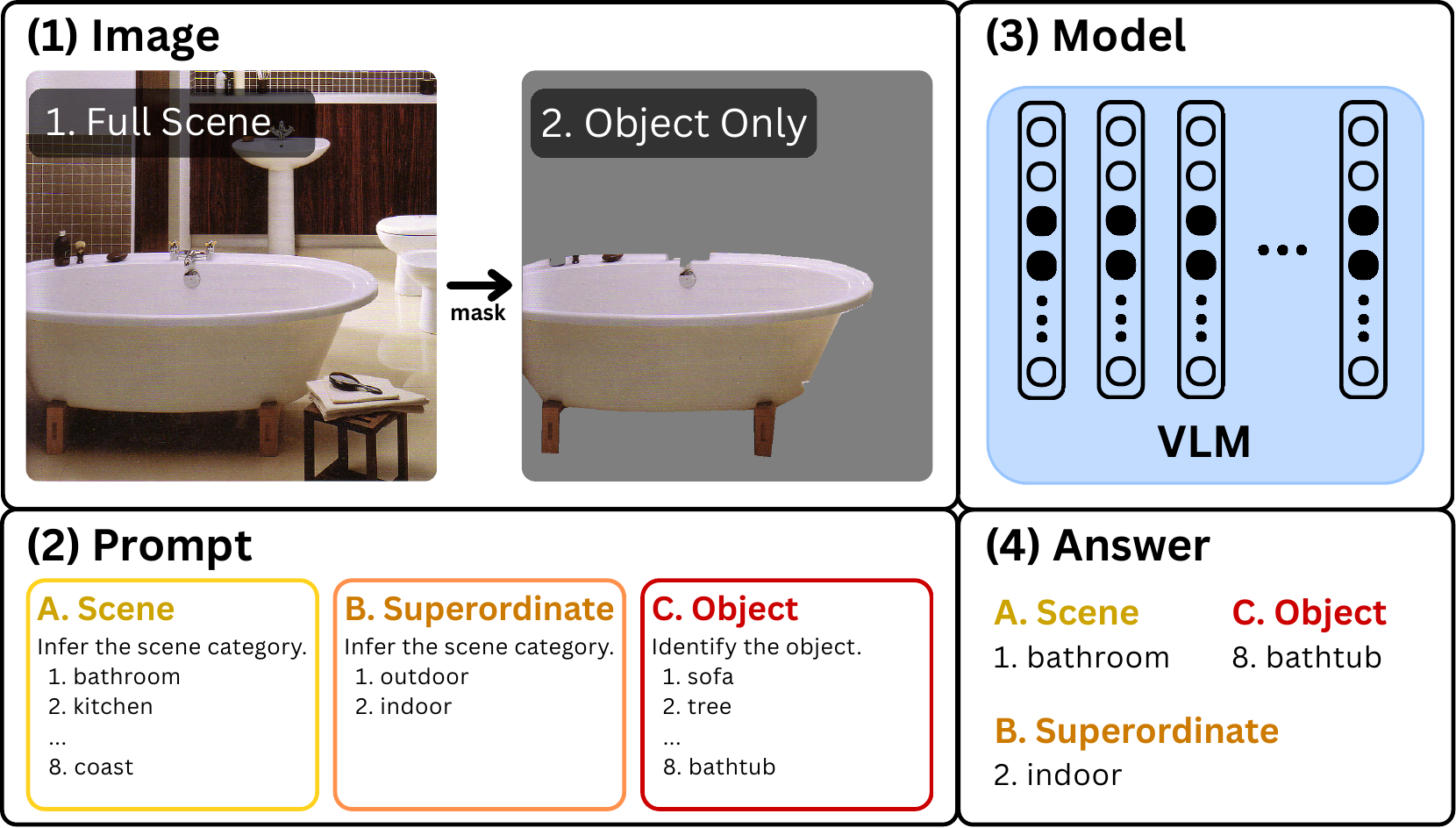}
  \vspace{-10pt}
  \caption{Experimental design. Each image condition (full scene or object-only) is paired with three prompt types: (A) scene category, (B) superordinate category, and (C) object identity. We analyze the intermediate image token representations corresponding to the objects (shown in black).}
  \vspace{-8pt}
  \label{fig:framework}
\end{wrapfigure}

A central question in scene perception research is whether contextual understanding is an emergent property of object representations. 
Cognitive studies have shown that humans can rapidly infer contextual information from single objects, and this capacity is modulated by object properties that reflect the statistical structure of object-scene relationships in natural images~\citep{wiesmann2023disentangling}. 
Whether the same holds for artificial multimodal systems remains largely unexplored.
Understanding how VLMs perform contextual inference from single objects is not only a theoretical question but one with direct practical consequences. 
It is essential for diagnosing model failures, assessing generalization, and building more robust VLMs.


We address this question through a behavioral and mechanistic investigation. 
We present two VLMs with images of single objects either embedded in their full scene or isolated on a masked background (see Figure \ref{fig:framework}). 
We probe the behavior and internal representations of these models across two levels of contextual abstraction -- scene category and superordinate category (indoor vs.\ outdoor) -- to ask:

\begin{itemize}
\item Do single objects carry scene-level and superordinate information independently of background context?
\item Is contextual inference modulated by the same object properties that predict human scene categorization?
\item Is contextual inference partially independent of object identification?
\item Are scene-level and superordinate schemas processed through the same internal pathways, or do they arise from distinct mechanisms?
\item Is contextual information encoded in object-patch representations, and how does the presence or absence of background context modulate these representations?
\end{itemize}

Our results show that single objects activate scene and superordinate schemas, with performance modulated by the same object properties that predict human scene categorization.
However, we found important distinctions in how this capacity is organized and grounded across tasks, object properties, and models.
Scene and superordinate inference draw on different object properties and are grounded differently in image token representations. Contextual inference is partially independent of object identification, and the two models differ substantially in how tightly object, scene, and superordinate information are integrated. 





\section{Related Work}


\paragraph{Contextual effects on object recognition in VLMs.}
The effect of contextual information on object recognition in VLMs has been previously studied.
\citet{rajaei2025} ask whether VLMs exhibit human-like contextual facilitation: if objects embedded in coherent scenes are recognized better than objects in phase-scrambled scenes, as they are for humans.
Similarly, \citet{li2025oric} show that contextual incongruity (objects appearing in unexpected scenes) systematically degrades object recognition in VLMs, with models hallucinating contextually expected but absent objects.
\citet{merlo2025cooco} examine how scene context around an object affects the model's language generation about that object, and uses attention patterns as a proxy for mechanistic insight. 
We go in the opposite direction: we remove the scene and ask what the object alone tells the model about its contect, and explain how this inference is grounded mechanistically.


\paragraph{Mechanistic Interpretability of VLMs.}
Previous work has found that image token representations in VLMs become progressively interpretable in the vocabulary space, with object information spatially localized in tokens corresponding to the object's image region \citep{neo2025, Schaumloeffel_2026_CVPR}.
We adopt the same technique to investigate how contextual information is grounded in object tokens when the scene background is absent.

\section{Methods}

\subsection{Models and Stimuli}

We evaluate two widely used large vision-language models (VLMs): LLaVA~1.5-13B 
\citep{liu2024llava} and InternVL3.5-14B \citep{wang2025internvl3_5} (see Appendix \ref{appendix_models} for details).

We use a curated subset of the dataset introduced by \citet{greene2013}, comprising $N = 2{,}392$ object-scene pairs drawn from $1{,}004$ unique images across eight scene categories (bathroom, bedroom, kitchen, living room, coast, forest, mountain, and skyline).
Each image contains multiple objects, but each instance selected for analysis focuses on a single 
foreground object for which a segmentation mask is available.
Full details on the instance selection procedure are provided in Appendix~\ref{appendix_dataset}.
For each object instance we record four scene-relevant properties previously investigated in human scene categorization research~\citep{wiesmann2023disentangling}:


\begin{itemize}
    \item \textbf{Frequency}: $P(\text{object}|\text{scene})$ -- the proportion of images within the associated scene category that contain the object, irrespective of how many times the object appears within a single image.
    \item \textbf{Specificity}: $P(\text{scene}|\text{object})$ -- the proportion of images containing the object that belong to the associated scene category, capturing how exclusively the object predicts that scene.
    \item \textbf{Size}: The fraction of the image area covered by the object mask.
    \item \textbf{Object Type}:  Following \citet{wiesmann2023disentangling}, we distinguish between two object types. Anchor objects are large, typically stationary objects that predict the presence and location of smaller objects in their vicinity (e.g., a stove predicting the presence of a pan). Local objects, by contrast, are the smaller objects whose position and identity are predicted by anchors (e.g., the pan). We extract the object type annotations from \citet{Turini2022}.
\end{itemize}

Frequency and specificity are computed using the ADE20K dataset \citep{ade20k}, as its substantially larger size provides more reliable estimates of object--scene co-occurrence statistics.

\subsection{Experimental Conditions}

Each image is presented to each VLM under two viewing conditions: 
\textbf{(1) Full scene}: the original unmodified image; 
\textbf{(2) Object only}: only the object is shown, with the rest of the image replaced by a grey background.

\subsection{Tasks}

The VLMs are queried with three successive forced-choice prompts, constructed independently per image. 
All prompts use an enumerated format in which answer options are listed with numerical indices, presented in randomized order.
Model responses were generated greedily, to ensure reproducibility. 
Full prompt templates are provided in Appendix~\ref{appendix_classif_task}.

The tasks are: \textbf{(1) Scene classification:} The model is asked to infer the scene category from one of the eight options. 
In the object-only condition, the prompt explicitly indicates that the background is masked, and the model must infer the scene from the visible object alone.
\textbf{(2) Superordinate classification:} The model classifies the scene as \emph{indoor} or \emph{outdoor}.
\textbf{(3) Object classification:} The model identifies the target object from eight options: the target object plus one distractor sampled from each of the remaining scene categories, drawn from a pre-compiled list of scene-typical objects. 

For the scene classification task, accuracy is scored in two ways. 
\emph{Normal accuracy} requires an exact match of the target scene label in the model response. 
\emph{Relaxed accuracy} counts a response as correct if any scene category associated with the queried object appears in the answer, where association is defined as the object having appeared in at least one image of that scene category in the dataset. 
This accounts for cases where the model correctly identifies a plausible scene that is not the ground-truth scene for the specific image but is semantically valid given the object.

\section{Experiments and Results}

\subsection{Single objects carry partial contextual information}
\label{sec:classification}

\begin{figure}[h]
    \centering
    \includegraphics[width=\textwidth]{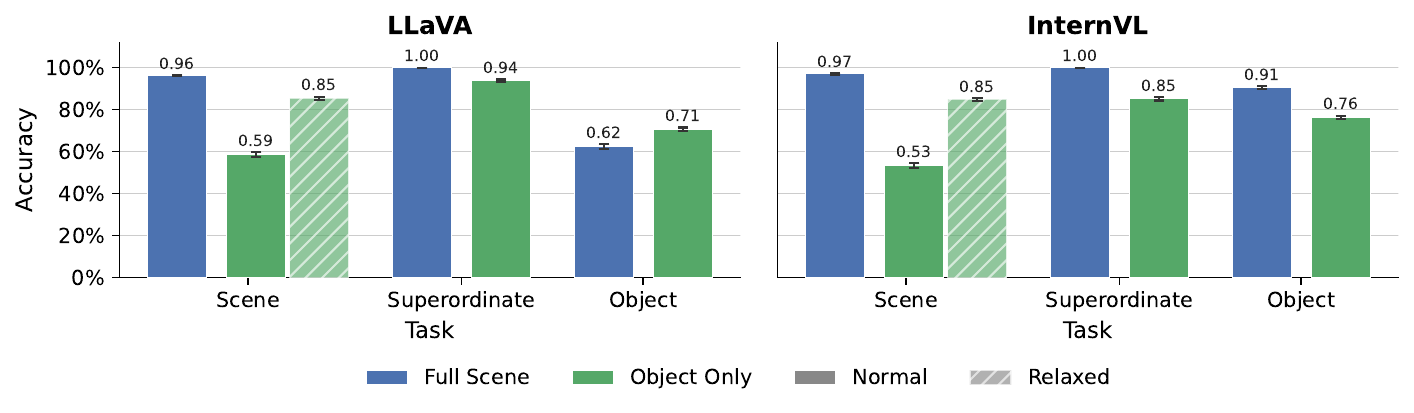}
    \vspace{-20pt}
    \caption{Classification accuracy across image conditions and tasks. Bars show accuracy under the full-scene (original image), and object-only (single foreground object on a masked background) conditions. For scene classification, solid bars indicate normal accuracy (exact label match) and hatched bars indicate relaxed accuracy (any semantically valid label accepted).  Error bars denote the standard error of the mean.}
    \label{fig:accuracy}
\end{figure}

As Figure \ref{fig:accuracy} shows, scene classification accuracy is near ceiling for both models under the full-scene condition (LLaVA\@: 96.2\%; InternVL\@: 97.0\%),  confirming that scene recognition is robust.
When the background is removed and the model must rely solely on the visible object, performance drops substantially (LLaVA\@: 58.6\%; InternVL\@: 53.4\% under normal scoring). 
However, the gap between normal and relaxed accuracy in the object-only condition suggests that models frequently retrieve a scene that is semantically consistent with the object but does not match the ground-truth scene  (LLaVA relaxed: 85.5\%; InternVL\@: 84.7\%). 
This pattern indicates that a single object activates plausible scene-level associations, but those associations are not sufficiently specific to identify the exact scene depicted.
We additionally find that scene classification accuracy in the object-only condition varies systematically by scene type (see Appendix~\ref{appendix_scene_acc}).
Outdoor scenes are consistently and substantially easier than indoor scenes for both models. 
The relaxed--normal accuracy gap is considerably larger for indoor scenes, reflecting greater category ambiguity.

Superordinate classification is markedly more robust than scene classification, with LLaVA reaching 93.8\% and InternVL 85.0\% in the object-only condition. 
The asymmetry between superordinate and scene-level accuracies mirrors findings from human scene categorization, where superordinate indoor/outdoor discrimination from objects also exceeds fine-grained scene-level  classification \citep{wiesmann2023disentangling}.

Object classification reveals a dissociation between the performance of the two models. 
In LLaVA, object recognition accuracy is higher when the scene background is absent than when it is present (full-scene: 62\%; object-only: 71\%), suggesting that background context introduces conflicting or distracting visual information that interferes with object identification in this model. 
InternVL shows the opposite pattern: accuracy drops when the full scene is unavailable (full-scene: 91\%; object-only: 76\%), although it still maintains a similar accuracy to LLaVA.
This suggests background information aids recognition of objects in this model.


\subsection{Contextual inference is modulated by object diagnostic properties}
\label{sec:diagnostic}

\begin{wrapfigure}{r}{0.63\textwidth}
  \vspace{-10pt}
  \centering
  \includegraphics[width=0.61\textwidth]{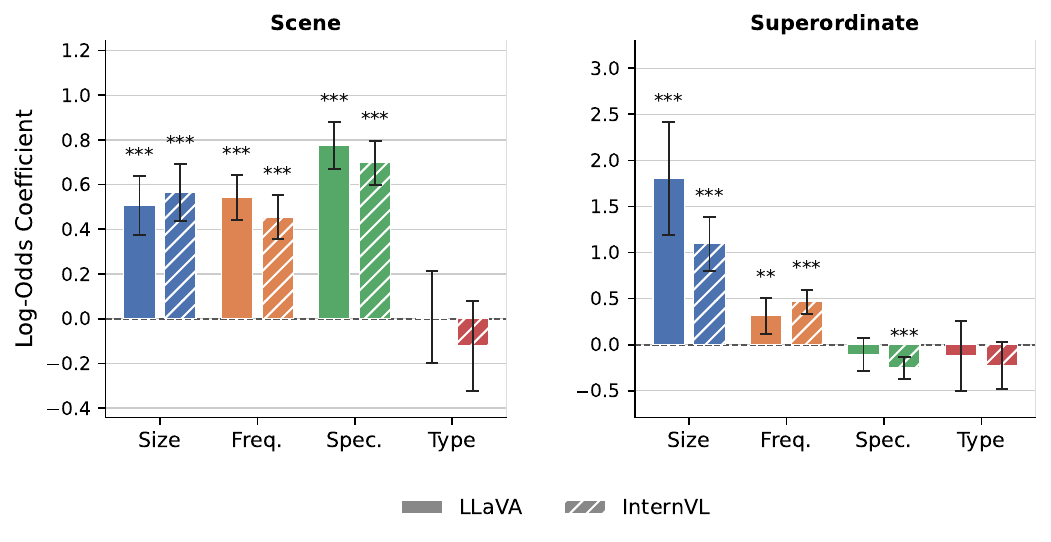}
  \vspace{-8pt}
  \caption{Log-odds coefficients from multivariate logistic regression predicting classification accuracy in the object-only condition, as a function of object size, frequency (Freq.), specificity (Spec.), and object type (anchor vs. local). Error bars denote 95\% confidence intervals. Significance levels: 
$^{**}p < .01$, $^{***}p < .001$.}
    \vspace{-15pt}
  \label{fig:regression}
\end{wrapfigure}

To identify which object properties drive scene and superordinate inference from single objects, we quantify the independent contribution of each property to classification accuracy in the object-only condition using multivariate logistic regression, following the approach of~\citet{wiesmann2023disentangling}:

\begin{equation}
\label{eq:logreg}
\log \frac{P(\text{correct})}{1 - P(\text{correct})} = \beta_0 
    + \beta_1 z(\text{frequency}) 
    + \beta_2 z(\text{specificity}) 
    + \beta_3 z(\text{size}) 
    + \beta_4 \mathbbm{1}[\text{type}] 
    + \varepsilon
\end{equation}

\noindent where $z(\cdot)$ denotes z-scoring (all continuous predictors are z-scored prior to fitting) and $\mathbbm{1}[\text{type}]$ is an indicator variable for object type (anchor vs.\ local). 
Separate regressions are fit for each task and model, and we report log-odds coefficients (Figure \ref{fig:regression}).

\paragraph{Scene classification.}
Object frequency, specificity, and size each independently and significantly predict scene classification accuracy in both models (all $p < .001$). 
Under normal scoring, specificity is the strongest  predictor in both LLaVA ($\hat{\beta}_\text{spec} = 0.78$) and InternVL  ($\hat{\beta}_\text{spec} = 0.70$), followed by frequency (LLaVA\@: $\hat{\beta}_\text{freq} = 0.54$; InternVL\@: $\hat{\beta}_\text{freq} = 0.45$)  and size (LLaVA\@: $\hat{\beta}_\text{size} = 0.51$; InternVL\@: 
$\hat{\beta}_\text{size} = 0.57$). 
Object type (anchor vs.\ local) does not significantly predict scene accuracy in either model once other continuous properties are controlled, suggesting that their distinction does not contribute independently beyond the statistical properties it correlates with.
This finding also mirrors that of humans reported in \citet{wiesmann2023disentangling}.


\paragraph{Superordinate classification.}
The predictor structure shifts markedly for the indoor/outdoor discrimination. 
Size is the dominant predictor in both models (LLaVA\@: $\hat{\beta}_\text{size} = 1.80$; InternVL\@: $\hat{\beta}_\text{size} = 1.09$, both $p < .001$), and  frequency contributes positively in both (LLaVA\@: $\hat{\beta}_\text{freq} = 0.31$, $p < .01$; InternVL\@: $\hat{\beta}_\text{freq} = 0.47$, $p < .001$). 
\sloppy
Specificity is non-significant for LLaVA but becomes a significant negative predictor for InternVL ($\hat{\beta}_\text{spec} = -0.25$, $p < .001$).
This suggests that objects highly diagnostic of a single fine-grained scene are actively disadvantageous for InternVL's superordinate categorization once size and frequency are controlled, possibly because such objects are strongly associated with a specific indoor or outdoor scene in a way that does not 
generalize to the broader category level. 
Object type does not significantly predict superordinate accuracy in either model.



\subsection{Contextual predictions are partially dissociable from object identity}

The analyses in Section~\ref{sec:classification} and Section~\ref{sec:diagnostic} 
establish that objects carry partial scene and superordinate information, and that 
this information is modulated by object properties. Here we ask a complementary 
question: when models fail, do their errors reveal a systematic relationship between 
object identity and scene-level inference? 
We examine this from two angles: whether correct object identification  is necessary for correct scene and superordinate inference (Section~\ref{sec:scene_inference}), and whether the model's scene and superordinate predictions are compatible with its object prediction (Section~\ref{sec:consistency}).

\subsubsection{Contextual inference does not require correct object identification}
\label{sec:scene_inference}

To assess whether scene and superordinate inference depend on correct object identification, we computed classification accuracy in the object-only trials conditioned on whether the model correctly identified the presented object.
We thus compare the scene and superordinate classification accuracy when the object was correctly identified,  when it was incorrectly identified, and the marginal (unconditioned) accuracy across all trials (see Table~\ref{tab:q2_accuracy}).

Correct object identification is associated with higher scene accuracy in both models (LLaVA\@: 65.6\% vs.\ 41.8\%; InternVL\@: 60.4\% vs.\ 31.0\%), under both normal and relaxed scoring. 
This association is stronger in InternVL, which shows a larger gap between object-correct and object-incorrect trials than LLaVA (29.4\% vs. 23.8\% under normal scoring; 24.9\% vs. 15.4\% under relaxed scoring). 
A similar pattern holds for superordinate classification, though accuracy is generally higher overall: correct object identification is associated with higher superordinate accuracy in both models (LLaVA\@: 96.0\% vs.\ 88.7\%; InternVL\@: 90.6\% vs.\ 67.1\%), and the gap is again larger in InternVL. 

Critically, the link is not perfect in either model: scene and superordinate accuracy remain well above zero even when the object is misidentified, and below ceiling even when it is correct. 
This indicates that scene and superordinate schemas are not strictly contingent on correct object identification: the models can retrieve scene-level information even when they fail to name the object, pointing to scene information being encoded in a partially independent representational pathway.


\begin{table}[h]
\centering
\caption{Scene and superordinate accuracy conditioned on object identification correctness in the object-only condition.}
\vspace{5pt}
\label{tab:q2_accuracy}
\setlength{\tabcolsep}{6pt}
\begin{tabular}{llccc}
\toprule
\textbf{Model} & \textbf{Task} & \textbf{Obj.\ correct} & \textbf{Obj.\ incorrect} & \textbf{Marginal} \\
\midrule
\multirow{3}{*}{LLaVA}
  & Scene           & 65.6\% {\color{green!60!black}(+7.0\%)}  & 41.8\% {\color{red!70!black}(-16.8\%)} & 58.6\% \\
  & Scene (relaxed) & 90.0\% {\color{green!60!black}(+4.5\%)}  & 74.6\% {\color{red!70!black}(-10.9\%)} & 85.5\% \\
  & Superordinate   & 96.0\% {\color{green!60!black}(+2.2\%)}  & 88.7\% {\color{red!70!black}(-5.1\%)}  & 93.8\% \\
\midrule
\multirow{3}{*}{InternVL}
  & Scene           & 60.4\% {\color{green!60!black}(+7.0\%)}  & 31.0\% {\color{red!70!black}(-22.4\%)} & 53.4\% \\
  & Scene (relaxed) & 90.6\% {\color{green!60!black}(+5.9\%)}  & 65.7\% {\color{red!70!black}(-19.0\%)} & 84.7\% \\
  & Superordinate   & 90.6\% {\color{green!60!black}(+5.6\%)}  & 67.1\% {\color{red!70!black}(-17.9\%)} & 85.0\% \\
\bottomrule
\end{tabular}
\end{table}

\subsubsection{Contextual and object predictions are not always compatible}
\label{sec:consistency}

To assess whether the model responds coherently across tasks, we assess whether the model's predicted scene and predicted superordinate are ones that can plausibly contain the predicted object in the same trial.
For example, if in the object classification task the model incorrectly predicted the presence of a bed, it should predict "bedroom" and "indoor" in the other classification tasks for the same trial.
Inconsistency indicates that the three inferences are not derived from the same underlying representation.

Both models show above-chance consistency. 
In LLaVA, the predicted superordinate is consistent with the predicted object in 71\% of trials, and the predicted scene is consistent with the predicted object in 62\% of trials. 
InternVL shows substantially higher consistency across both measures (superordinate: 90\%; scene: 84\%), indicating that its scene and superordinate predictions more often correspond to plausible contexts for the object it identified. 
The gap between the two models is considerable: InternVL's scene and superordinate predictions are compatible with its object prediction roughly 20\% more often than LLaVA's. 

Consistency is nonetheless imperfect even in InternVL, indicating that the predicted scene and superordinate are not always compatible with the predicted object. 
This residual inconsistency, combined with the partial dissociation between object recognition accuracy and scene inference reported in Section~\ref{sec:scene_inference}, suggests that scene and superordinate predictions are not strictly derived from the object prediction alone, and different pathways are used to answer the tasks.

\subsection{Contextual inference is grounded in object-patch representations}
\label{sec:grounding}

The analyses so far characterize \textit{what} information objects carry about scenes and superordinate categories, and how errors in contextual inference relate to object identity. 
We now turn to the question of \textit{how} this information is grounded mechanistically: where in the network does scene and superordinate information reside, and how does the presence or absence of background context modulate the object's internal representation? We address this through two complementary analyses. Section~\ref{sec:stability} examines representational stability (the degree to which object-patch hidden states change when background context is removed) and tests whether this stability predicts classification accuracy across layers. 
Section~\ref{sec:logit} then asks whether scene and superordinate information is directly decodable from the image vocabulary space, and at which processing depth it emerges.

\subsubsection{Representational stability predicts contextual inference accuracy}
\label{sec:stability}

The behavioral results establish that objects carry partial scene and superordinate information, and that this is modulated by object properties. 
A natural mechanistic question follows: does the object's internal representation actually change when background context is removed, and if so, does the degree of that change predict whether scene and superordinate inference succeeds? 
An object whose representation is robust to context removal, meaning one whose hidden-state activations are similar whether or not the background is present, can be said to carry self-sufficient scene information intrinsically, independent of contextual support. 
Conversely, an object whose representation shifts substantially when background is removed relies more heavily on contextual integration with the object representation.
We operationalize this notion as \textit{representational stability} and test whether it predicts classification accuracy across layers and tasks.

To assess whether object representations are modulated by background context, we computed the cosine similarity between hidden-state activations of object-patch tokens under the full-scene and object-only conditions across all transformer layers:

\begin{equation}
\label{eq:cossim}
\text{CosSim}^{(\ell)}_i = \frac{1}{|P_i|} \sum_{p \in P_i} 
\cos\left(h^{(\ell)}_{i,p,\text{full}},\ h^{(\ell)}_{i,p,\text{object}}\right)
\end{equation}

\noindent where $P_i$ is the set of patch indices corresponding to the foreground object of image $i$, determined by projecting the foreground mask onto the image patch grid and retaining only patches fully covered by the object mask, and $h^{(\ell)}_{i,p}$ is the hidden state of patch $p$ at layer $\ell$. 
A high cosine similarity indicates that the object's representation is stable regardless of 
whether the surrounding scene context is present.

\begin{figure}[h]
    \centering
    \includegraphics[width=\textwidth]{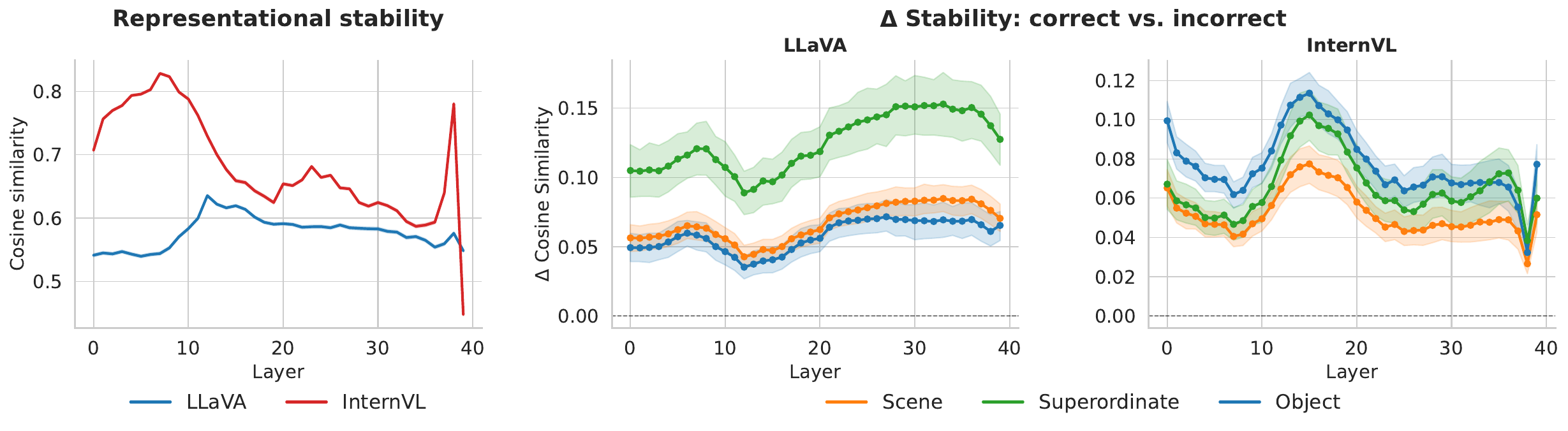}
    \vspace{-10pt}
    \caption{Representational stability of object-patch tokens across transformer layers. \textit{Left:} Mean cosine similarity between hidden-state activations of object patches under the full-scene and object-only conditions, averaged across all images, for LLaVA and InternVL. Higher values indicate that object representations are less modulated by the presence of background context. \textit{Center and right:} Difference in mean cosine similarity between correctly and incorrectly classified trials ($\Delta$ cosine similarity) at each layer. Values above zero indicate that correctly classified trials have more stable object representations. Shaded regions denote 95\% confidence 
intervals.}
    \label{fig:repr_stability}
\end{figure}

InternVL shows higher representational stability overall than LLaVA (mean cosine similarity: LLaVA $= 0.57$; InternVL $= 0.68$), indicating that its object-patch representations are less modulated by the presence or absence of background context. 
The two models also show distinct layer-wise profiles: InternVL's cosine similarity decreases across layers, while LLaVA's increases
(Figure~\ref{fig:repr_stability}; first panel).

To determine whether representational stability predicts classification accuracy, we computed at each layer the difference in mean cosine similarity between correctly and incorrectly classified trials, assessing statistical significance via a two-sided permutation test (1,000 permutations) in which trial labels were randomly shuffled to generate a null distribution of mean differences. 
Correctly classified trials show higher representational stability than incorrectly classified trials across all tasks and both models (Figure~\ref{fig:repr_stability}; second panel): objects whose representations are robust to context removal carry more self-sufficient scene and superordinate information. 
The layer at which this difference peaks differs between models: in LLaVA the peak occurs in later layers, consistent with context-dependence being primarily a late-layer phenomenon in this model. 
In InternVL the peak occurs in middle layers, where the overall decrease in cosine similarity across layers is most pronounced, suggesting that the processing stages at which object representations are most sensitive to context removal are also those most critical for correct scene and superordinate inference

Notably, object identification accuracy is also positively associated with representational stability, particularly in InternVL, suggesting that context-independent object representations benefit not only scene-level inference but object recognition itself.

Since cosine similarity is averaged over object patches, objects occupying a larger image area contribute more patches and could yield more reliable estimates by construction. 
To rule out this confound, we included z-scored object size as a covariate in a logistic regression predicting task accuracy from z-scored cosine similarity. Cosine similarity remained a significant positive predictor of performance across all tasks and both models after controlling for object size (all $p < .001$; see Table~\ref{tab:regression_cosine_size} in Appendix).

\subsection{Contextual information emerges differently across tasks and models}
\label{sec:logit}

Prior work has shown that semantic information is grounded in image tokens and can be decoded using output vocabulary projection methods~\citep{neo2025, Schaumloeffel_2026_CVPR}. 
Here, we use this technique to ask the following question: does this scene and superordinate information emerge directly in the image-patch-level vocabulary space, and if so, at which processing depth?

For each image in the object-only condition, we project image patch representations at each layer to the vocabulary space using the unembedding matrix, and identify the three patches with the highest logit value for the correct scene or superordinate label. 
We then compute the mean logit across these top-3 patches as a layer-wise measure of how strongly the correct label is encoded in image tokens.
To test whether logit strength continuously predicts classification accuracy, we computed the ROC-AUC between the mean logits and binary accuracy at each layer.
Statistical significance was assessed via a one-sided Mann-Whitney U test comparing logit strength between correctly and incorrectly classified trials at each layer.

\begin{wrapfigure}{r}{0.6\textwidth}
  \vspace{-10pt}
  \centering
  \includegraphics[width=0.58\textwidth]{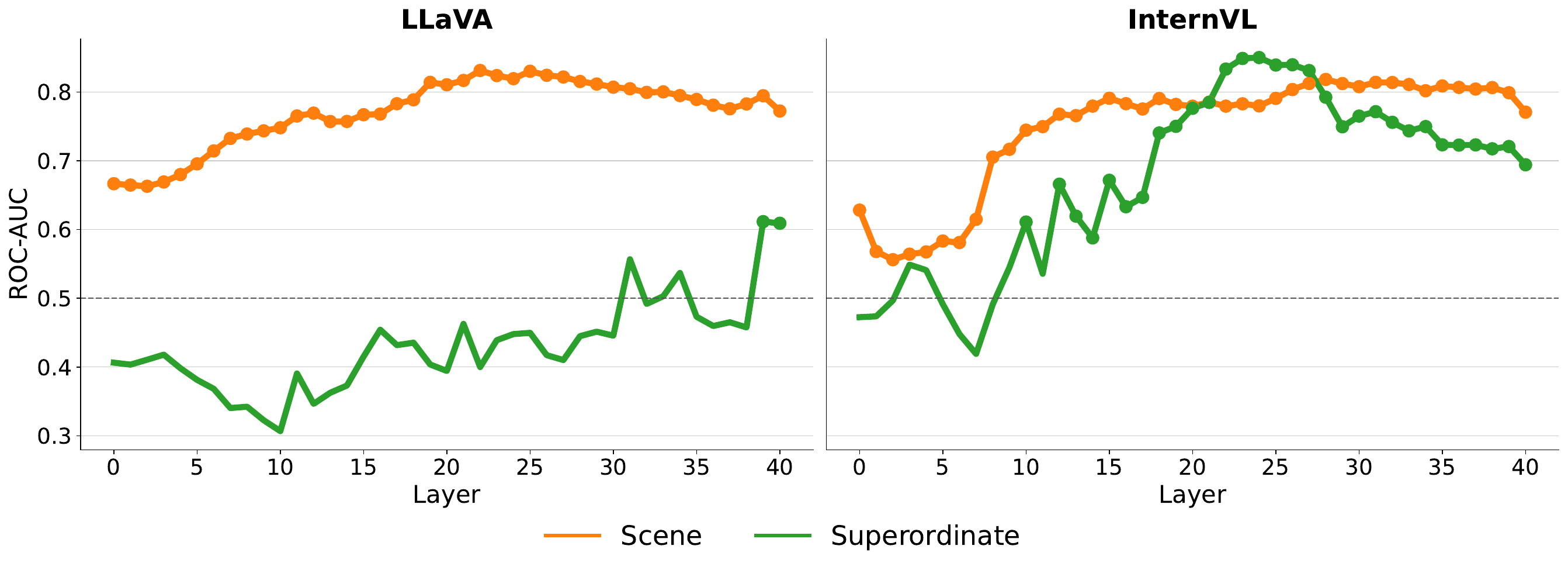}
  \vspace{-10pt}
  \caption{ROC-AUC between top-3 patch logit strength and binary classification accuracy at each transformer layer. Values above the dashed line indicate that logit strength in image patch tokens is predictive of classification accuracy at that layer. Markers indicate statistically significant layers.}
  \vspace{-8pt}
  \label{fig:logit}
\end{wrapfigure}

We found that scene logit strength is a strong and consistent predictor of scene classification 
accuracy in both models from the very first transformer layer (LLaVA\@: $AUC = 
0.67$; InternVL\@: $AUC = 0.63$), increasing to a peak of approximately $0.81$ 
and $0.79$ respectively in mid-to-late layers (Figure~\ref{fig:logit}). 
The signal is significant across nearly all layers in both models, indicating that scene-discriminative information is present in image tokens throughout the entire depth of the network.

Superordinate accuracy tells a different story. In LLaVA, superordinate AUC 
remains below chance for the majority of the network, only crossing chance level in 
the final few layers, indicating that the vocabulary projection of image tokens 
is not predictive of superordinate categorization for most of the network's 
depth. In InternVL, superordinate AUC begins near chance, then rises steadily to $0.85$ around layers 25--28, eventually exceeding 
scene AUC at its peak before declining. Superordinate information thus emerges 
in image tokens in InternVL, but later than scene information.

These results reveal a fundamental dissociation in how scene and superordinate schemas are grounded in image tokens. 
Scene identity is encoded in the patch-level vocabulary space from the earliest layers and persists throughout processing in both models. 
Superordinate categorization, by contrast, is not meaningfully encoded in this space in LLaVA at any depth, suggesting it is computed through a different mechanism.
In InternVL, superordinate information does eventually emerge in image tokens, but through a slower, depth-dependent process distinct from scene encoding. 
Together, these findings provide mechanistic evidence that scene and superordinate schemas are not co-localized in image token  representations, and that model architecture critically determines whether and  where superordinate information becomes accessible in the visual processing stream of the VLM.

\section{Discussion}


A central question in scene perception research is whether scene understanding is an emergent property of object representations, or whether it requires holistic scene-level processing that cannot be reduced to individual objects. 
Work in human psychophysics has established that single objects can support above-chance scene categorization and that this capacity is modulated by the same statistical properties that predict object-scene co-occurrence in natural images~\citep{wiesmann2023disentangling}. 
Our results show that VLMs exhibit  a qualitatively similar capacity: single objects activate scene-level and superordinate schemas above chance, with performance modulated by the same object properties as humans. 
This convergence suggests that the statistical structure of natural scene-object relationships is internalized during training of VLMs in a way that recapitulates key properties of human perceptual learning.


In addition, our findings show that scene inference is partially independent of object identification: scene and superordinate accuracy remain well above zero even when the model fails to identify the object, and scene and superordinate predictions are not always compatible with the object prediction even in InternVL, which shows the highest internal consistency. 
This suggests that contextual inference draws at least partially on representational pathways that are distinct from those underlying object identification.

We also found evidence that scene and superordinate classification involve different processing. 
Behaviorally, scene and superordinate performance differ. 
Moreover, scene accuracy is driven by specificity and frequency, while superordinate is dominated by object size. 
The dependence on correct object identification also differs: correctly identifying the object yields a larger boost for scene than for superordinate classification in both models. 
Mechanistically, scene identity is encoded in image tokens from the first layer throughout the network, while superordinate information is absent from LLaVA's image tokens at any depth and emerges only late in InternVL. 

Finally, the two models differ in how tightly all of these levels are integrated, and this difference is visible across every analysis. 
InternVL shows stronger coupling between object identification and contextual inference, higher behavioral consistency across tasks, higher representational stability, and evidence of both scene and superordinate being grounded in image tokens. 
LLaVA processes the three tasks with greater modularity: its object recognition is disrupted by scene context, its predictions are less internally coherent, its representational stability is more brittle to contextual removal, and its superordinate representations are not grounded in the vocabulary space of image tokens.
Critically, these differences form a coherent pattern. 
A model that binds object, scene, and superordinate information into a more integrated representational structure makes more internally coherent errors, shows stronger behavioral coupling across tasks, and grounds more of its predictions in visual rather than purely linguistic processing. 
Understanding what drives this difference in integration, whether through architecture, pretraining data, or instruction-tuning, is a key direction for future work, and may illuminate what computational conditions are necessary for the kind of integrated scene-object processing.

\subsection{Limitations}

In this work, we evaluate a fixed set of eight scene categories; while these were selected to be representative of both indoor and outdoor environments, generalizability to a broader or more fine-grained taxonomy remains to be tested. 
Second, we evaluate two models; while LLaVA and InternVL were chosen as widely used representatives that remain architecturally distinct, the degree to which the results generalize across other model families warrants further investigation. 
Finally, while our findings parallel human scene categorization in several respects, a direct comparison with human behavioral data on the same stimuli would be needed to quantify the degree of alignment more precisely, and would open the door to testing the speculative predictions about representational entanglement raised in this work.
Similarly, further mechanistic work is needed to characterize the nature of the distinct internal representations underlying object identification, scene inference, and superordinate categorization within the network.

\bibliography{iclr2026_conference}
\bibliographystyle{iclr2026_conference}

\appendix

\section{Model Details}
\label{appendix_models}

We compare two vision-language models (VLMs): LLaVA~1.5-13B~\citep{liu2024llava}
(\texttt{llava-hf/llava-1.5-13b-hf}) and InternVL3.5-14B~\citep{wang2025internvl3_5} 
(\texttt{OpenGVLab/InternVL3\_5-14B-HF}), both accessed via the HuggingFace Transformers
library.
 
\paragraph{LLaVA-1.5-13B.}
LLaVA connects a visual backbone to a large language model (LLM) via a two-layer MLP
projection. Input images are padded to a square aspect ratio and resized to
$336 \times 336$ pixels. The visual backbone produces $24 \times 24$ patch embeddings,
which are projected into the LLM embedding space, yielding $576$ visual tokens per image.
 
\paragraph{InternVL3.5-14B.}
InternVL follows a similar architectural principle but extends LLaVA in two key ways.
First, the visual encoder accepts higher-resolution $448 \times 448$ pixel inputs.
Second, each $2 \times 2$ block of visual tokens is compressed into a single token via
pixel shuffling before being forwarded to the LLM, reducing the spatial grid from
$32 \times 32$ to $16 \times 16$ tokens (256 tokens total).
InternVL also supports dynamic high-resolution processing, in which an image is split
into multiple sub-images that are each encoded independently. Since our inputs are
generally small, we restrict this to a single view for all experiments.
 

\section{Dataset Preprocessing}
\label{appendix_dataset}

We apply five sequential filtering steps to obtain a clean and balanced set of object--scene pairs (with corresponding segmentation masks) from \citet{greene2013}:

\begin{enumerate}    
    \item \textbf{Scene category selection}. Only annotations belonging to eight scene categories (bathroom, bedroom, kitchen, living room, coast, forest, mountain, skyline) are retained. These categories provide a representative spread of both indoor and outdoor environments with high between-category discriminability.
    \item \textbf{Occlusion removal}. Images in which any anchor object's segmentation mask is covered by $\geq$50\% of surrounding local object masks are discarded. This ensures that anchor objects are not heavily occluded and remain recognizable.
    \item \textbf{Rare object removal}. Object categories appearing in fewer than 10 images are discarded. This avoids including object categories that are too infrequent in natural scene statistics.
    \item \textbf{Minimum size threshold}. Objects whose segmentation mask covers $\leq$3\% of the image area are removed, as objects of this size provide insufficient visual information for reliable recognition \citep{huang2017speedaccuracy}.
    \item \textbf{Balanced per-category subsampling}. For each scene category, up to 150 anchor objects and 150 local objects are randomly sampled, ensuring comparable sample sizes across categories.
\end{enumerate}

\section{Classification Task Details}
\label{appendix_classif_task}

Each image was presented to the model with three successive forced-choice prompts, constructed independently per image. All prompts used an enumerated format, in which answer options were listed with numerical indices and the model was instructed to respond in the exact format \texttt{[number]. [category name]}.

\paragraph{Scene classification prompt.}
The eight scene category labels were randomly shuffled on each trial and presented as a numbered list. The full prompt read:

\begin{quote}
\textit{Infer the scene category from the image. Available categories:}\\
\textit{1. [category$_{\sigma(1)}$]}\\
\textit{\hspace{1em}\vdots}\\
\textit{8. [category$_{\sigma(8)}$]}\\[4pt]
\textit{Respond in this exact format: [number]. [category name]}
\end{quote}

\noindent For the object-only condition, the preamble was replaced with: \textit{``The image shows a segmented object with the background masked in gray. Infer the scene category from the presented object.''} A response was scored as correct if either the ground-truth category name or its assigned index appeared in the model's output.

\paragraph{Superordinate classification prompt.}
The two options (\textit{indoor}, \textit{outdoor}) were randomly shuffled on each trial and presented in the same enumerated format:

\begin{quote}
\textit{Infer the scene category from the image. Available categories:}\\
\textit{1. [indoor \textbf{or} outdoor]}\\
\textit{2. [outdoor \textbf{or} indoor]}\\[4pt]
\textit{Respond in this exact format: [number]. [category name]}
\end{quote}

\paragraph{Object classification prompt.}
The candidate set consisted of the target object plus one distractor randomly sampled from each of the remaining scene categories, drawn from a pre-compiled list of scene-typical objects. This yielded $N_\text{scenes}$ options in total, which were randomly shuffled and presented in enumerated format:

\begin{quote}
\textit{Identify the object category present in the image. Available categories:}\\
\textit{1. [object$_{\sigma(1)}$]}\\
\textit{\hspace{1em}\vdots}\\
\textit{$N$. [object$_{\sigma(N)}$]}\\[4pt]
\textit{Respond in this exact format: [number]. [category name]}
\end{quote}

\noindent For the object-only condition the preamble read: \textit{``The image shows a segmented object with the background masked in gray. Identify the object category present in the image.''}

\paragraph{Implementation details.}
All random shuffling and distractor sampling used a fixed global random seed ($\text{seed} = 42$), set once before the dataset loop, ensuring reproducibility across runs. Model responses were generated greedily.

\section{Scene Classification Accuracy by Scene Type}
\label{appendix_scene_acc}

\begin{figure}[H]
    \centering
    \includegraphics[width=\textwidth]{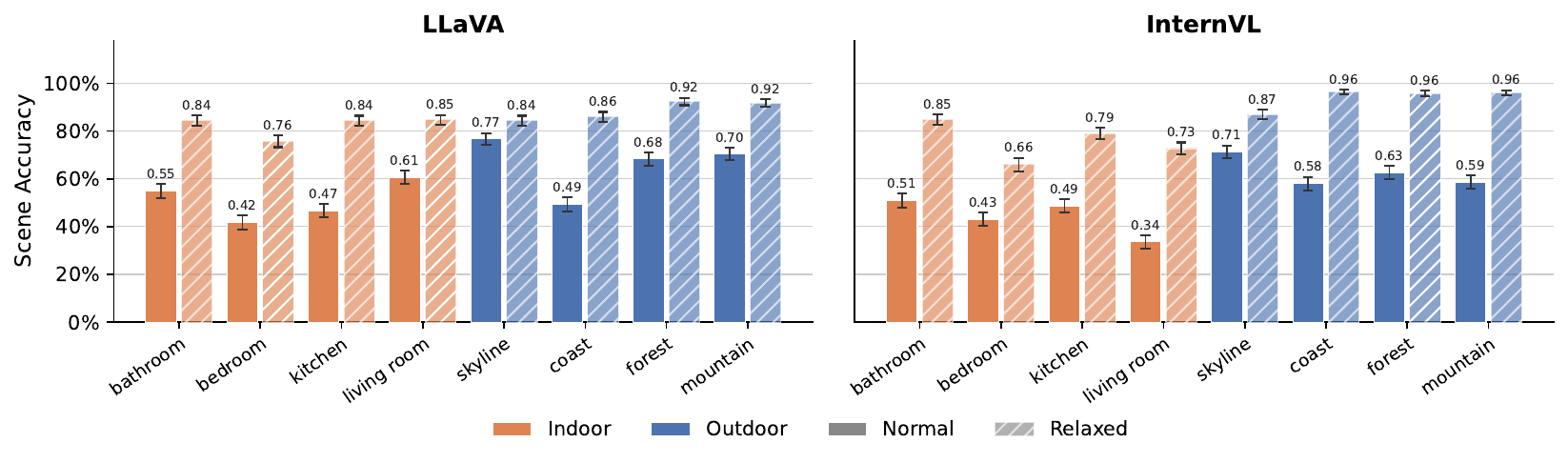}
    \label{fig:accuracy_scene}
    \caption{Scene classification accuracy in the object-only condition by scene category, for LLaVA and InternVL. Solid bars indicate normal accuracy (exact label match) and hatched bars indicate relaxed accuracy (any semantically valid label accepted). Orange bars correspond to indoor scene categories (bathroom, bedroom, kitchen, living room) and blue bars to outdoor categories (skyline, coast, forest, mountain). Error bars denote the standard error of the mean.}
\end{figure}

\subsubsection{Contextual inference from single objects has potential for improvement}


To quantify how much room exists for improving object-only contextual inference in VLMs, we establish a reference by measuring how well scene and superordinate categories can be inferred from the object label alone. 
We compare two variants: a mean-token baseline, which replaces the image with a semantically uninformative average visual token while providing the object name in the prompt, and a pure LLM baseline, which uses the LLM backbone of the VLM model that was trained exclusively on language data. For LLaVA this is Vicuna-13B~\citep{vicuna2023} (\texttt{lmsys/vicuna-13b-v1.5}), and
for InternVL this is Qwen3-14B~\citep{yang2025qwen3technicalreport} (\texttt{Qwen/Qwen3-14B}).

As prompts for the baselines, the object name is inserted into the prompt in place of the image, with scene and superordinate classification prompts otherwise unchanged. Thus, the prompt takes the form of:
\textit{The image shows a [object].}
\textit{[scene classification prompt \textbf{or} superordinate classification prompt]}.

For scene classification, relaxed scoring provides the more meaningful comparison: because a single object label is naturally consistent with multiple scene categories, normal scoring penalizes valid associations and systematically underestimates label-based inference. Under relaxed scoring, both models fall short of the mean-token baseline (LLaVA: 0.85 vs. 0.90; InternVL: 0.85 vs. 0.95). Relative to each model's own LLM baseline, the pattern differs: LLaVA exceeds it (0.85 vs. 0.79), indicating that the visual token adds scene-associative information beyond what LLaVA's language backbone recovers from the label alone, while InternVL falls below it (0.85 vs. 0.94), indicating that InternVL's visual token fails to match what its stronger language backbone infers from the label. For superordinate classification, both models fall below at least one baseline (LLaVA: 0.94 vs. MT 0.95, LLM 0.99; InternVL: 0.85 vs. MT 0.94, LLM 0.93), with InternVL again showing the largest gap. Together, these results indicate that current VLMs leave meaningful performance on the table in object-only contextual inference, pointing to substantial room for improvement.

\begin{figure}[H]
    \centering
    \includegraphics[width=\textwidth]{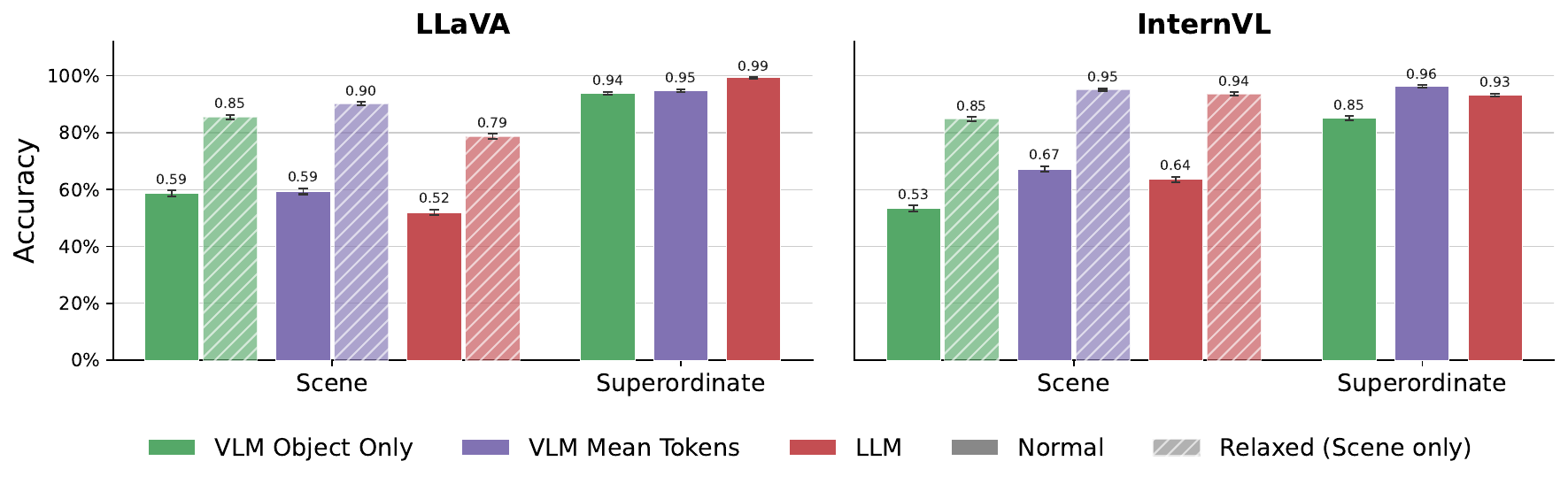}
    \label{fig:acc_baselines}
    \caption{Scene and superordinate classification accuracy for the object-only condition compared to label-based baselines. VLM Object Only shows VLM performance when only the foreground object is visible. VLM Mean Tokens replaces the image with a semantically uninformative average visual token while providing the object name in the prompt. LLM receives the object name but no image. For scene classification, solid bars indicate normal accuracy and hatched bars indicate relaxed accuracy. Superordinate classification uses normal scoring only. Error bars denote the standard error of the mean.}
\end{figure}

\section{Object-size control in representational stability}

\begin{table}[H]
\centering
\caption{Logistic regression predicting task accuracy from cosine similarity and object size
         (both z-scored). Coefficients are log-odds. Object size is included as a control
         for the number of patches contributing to the cosine similarity estimate.}
\label{tab:regression_cosine_size}
\small
\begin{tabular}{llrrrrl}
\toprule
\textbf{Model} & \textbf{Task} & \textbf{Predictor} & $\hat{\beta}$ & \textbf{SE} & $z$ & $p$ \\
\midrule
\multirow{6}{*}{LLaVA}
 & \multirow{2}{*}{Scene}
   & Cosine sim.  &  0.487 & 0.051 &  9.54 & $<.001$ \\
 & & Object size  &  0.385 & 0.061 &  6.30 & $<.001$ \\
\cmidrule{2-7}
 & \multirow{2}{*}{Superordinate}
   & Cosine sim.  &  1.017 & 0.109 &  9.36 & $<.001$ \\
 & & Object size  &  1.344 & 0.302 &  4.46 & $<.001$ \\
\cmidrule{2-7}
 & \multirow{2}{*}{Object}
   & Cosine sim.  &  0.728 & 0.058 & 12.57 & $<.001$ \\
 & & Object size  & -0.373 & 0.056 & -6.71 & $<.001$ \\
\midrule
\multirow{6}{*}{InternVL}
 & \multirow{2}{*}{Scene}
   & Cosine sim.  &  0.465 & 0.051 &  9.06 & $<.001$ \\
 & & Object size  &  0.443 & 0.058 &  7.59 & $<.001$ \\
\cmidrule{2-7}
 & \multirow{2}{*}{Superordinate}
   & Cosine sim.  &  0.567 & 0.069 &  8.16 & $<.001$ \\
 & & Object size  &  0.916 & 0.141 &  6.51 & $<.001$ \\
\cmidrule{2-7}
 & \multirow{2}{*}{Object}
   & Cosine sim.  &  0.910 & 0.066 & 13.73 & $<.001$ \\
 & & Object size  &  0.161 & 0.080 &  2.01 & $.044$  \\
\bottomrule
\end{tabular}
\end{table}



\end{document}